\documentclass[runningheads]{llncs}
\usepackage[T1]{fontenc}
\usepackage{graphicx}
\usepackage{cite}
\usepackage{amsmath,amssymb,amsfonts}
\usepackage{algorithmic}
\usepackage{graphicx}
\usepackage{textcomp}
\usepackage{xcolor}
\usepackage{float}
\usepackage[hidelinks]{hyperref}
\usepackage{xurl}
\usepackage{multirow}

\usepackage{orcidlink}
\usepackage{cite}

\begin{document}
\title{AI-Generated Image Recognition \\via Fusion of CNNs and Vision Transformers}
\titlerunning{AI-Generated Image Recognition}
%
\authorrunning{X.-B. Mai et al.}

\author{
Xuan-Bach Mai\inst{1,2}  \and
Hoang-Minh Nguyen-Huu\inst{1,2} \and
Quoc-Nghia Nguyen\inst{1,2} \and
Hoang-Tung Vu\inst{1,2} \and 
Minh-Triet Tran\inst{1,2}\orcidlink{0000-0003-3046-3041} \and
Trung-Nghia Le\thanks{Corresponding author.}\inst{1,2}\orcidlink{0000-0002-7363-2610}}

\institute{University of Science, VNU-HCM, Ho Chi Minh City, Vietnam \and
Vietnam National University, Ho Chi Minh City, Vietnam \\
\email{\{mxbach22,nhhminh22,nqnghia22,vhtung22\}@apcs.fitus.edu.vn, \{lkduy,tmtriet,ltnghia\}@fit.hcmus.edu.vn}
}

\maketitle              
\begin{abstract}

Recent advancements in synthetic data technology have opened a new era where images of remarkable quality are generated, blurring the lines between real-life images and those produced by Artificial Intelligence (AI). This evolution poses a significant challenge to ensuring the reliability and authenticity of data, underscoring the need for robust detection methods. In this paper, we present a robust approach aimed at addressing these pressing concerns. Our methodology revolves around leveraging fusion strategies, combining the strengths of multiple detection methods for identifying AI-generated images. Through extensive experimentation on the CIFAKE dataset, our model showcases remarkable performance, achieving an impressive accuracy rate of 97.32\%. This accomplishment underscores the efficacy of our approach in accurately distinguishing between AI-generated images and real-life images, thus contributing to the advancement of data authentication techniques amidst the proliferation of synthetic data.

\end{abstract}
\section{Introduction}

Artificial intelligence (AI)-generated images have become increasingly popular in recent years, with various tools and platforms available for users to create captivating visuals for social media and other purposes. One of the top recent image generators is DALL·E 3, known for its ability to produce high-quality images quickly \cite{openaiDALLE}\cite{zapierBestImage}. AI has the capacity to create images from scratch using trained artificial neural networks \cite{altexsoftImageGeneration}. This technology allows for limitless creativity at the fingertips of users, with AI engines capable of producing art that rivals human creativity\cite{wiredWhatAIGenerated}. Overall, AI-generated images have the potential to be of excellent quality and offer users a range of creative possibilities. As technology continues to advance, AI image generators are likely to become even more sophisticated and user-friendly, providing new opportunities for creativity and visual expression.

Detecting AI-generated images is crucial due to the potential harm they can cause in society. These images can be easily manipulated and altered to create misleading content, disseminating false stories and undermining the credibility of media sources. Moreover, the unauthorized use of personal images, such as those obtained from social media, poses significant privacy risks and potential security threats. Identity theft is another serious concern, as AI-generated images can be leveraged to fabricate convincing fake identities, putting individuals' reputations and online safety at risk. Furthermore, the non-consensual use of likenesses in AI-generated images raises ethical questions regarding ownership and control over one's identity. This technology enables the creation of images that closely resemble real people without their consent, highlighting the moral implications of using someone's likeness without permission. Additionally, the amplification of biases is a concerning issue, as AI algorithms trained on biased datasets can perpetuate and exacerbate existing societal inequalities and discriminatory practices. Ensuring that AI algorithms are trained on diverse and unbiased datasets is essential to mitigate these risks and promote fairness and inclusivity in AI-generated content\cite{aiornotAIGenerationPhotos}.

In the realm of AI, the detection of AI-generated images has become a crucial area of focus. Various tools and methods have been developed to identify whether an image has been created using AI technology. One popular option is the "Hive AI Detector," a Chrome extension that provides a score ranking the likelihood of an image being real or AI-generated\cite{aiornotDetectorCheck}. Additionally, AI detection tools allow users to upload an image and click a "Check" button to determine if it is AI-generated\cite{isitai}. When examining AI-generated images, it is important to look for inconsistencies that may indicate artificial intelligence was used in the creation process\cite{pcmagDetectAIGenerated}. Furthermore, industry-standard indicators have been developed to label AI-generated images on social media platforms like Facebook and Instagram\cite{fbLabelingAIGenerated}. These indicators help users identify images that have been created using AI technology. Another method to detect AI-generated images is by checking the image metadata, as this can reveal telltale signs that may not be visible to the naked eye\cite{forbesNotDetect}. Additionally, APIs like Illuminarty's AI detection tool can be integrated into services to automatically identify whether an image was generated using AI\cite{illuminartyGeneratedContent}. Despite the advancements in AI detection tools, there are still challenges in fooling these systems. Tools like Art Detector are designed to identify subtle markers embedded in AI-generated images, looking for unusual patterns that may indicate artificial intelligence was used in the creation process \cite{nytimesEasyFool}. Overall, the development of AI image detectors and detection methods continues to evolve as the use of AI technology in image creation becomes more prevalent\cite{huggingfaceImageDetector}.

Recognizing the crucial need for detecting AI-generated images, we propose a robust detection method to address the potential harms associated with digitally manipulated visuals. Our  approach revolves around leveraging fusion strategies to significantly boost the accuracy of our AI-generated image recognition model. Specifically, we combine the strengths of Convolutional Neural Networks (CNNs) and Vision Transformers (ViTs) through fusion algorithms. Extensive experiments on the CIFAKE dataset \cite{CIFAKE} shows that our proposed method achieve an impressive accuracy of 97.32\%.

Our main contribution lies in running individual detection models and fusing their outputs to create higher-accuracy models. This approach enhances detection precision and ensures our model's adaptability against a diverse range of AI-generated images. By integrating the capabilities of CNNs and ViTs, our goal is to build a detection system that effectively safeguards against the spread of misleading or harmful content in the digital realm.

In the remaining sections of this paper, we cover related methods, methodology, experimental results, conclusions, and future work. We discuss existing approaches to detecting AI-generated images in Section \ref{sec:related_work}, detail our methodology for developing our AI-generated image recognition model in Section \ref{sec:proposed_method}, present the experimental results obtained from testing our model, draw conclusions based on our findings in Section \ref{sec:experiments}, and outline future research directions in Section \ref{sec:conclusion}.

\section{Related Work}
\label{sec:related_work}

\subsection{Generative Models}

Generative models constitute a pivotal domain within artificial intelligence, aiming to capture and model the inherent distribution of data from observed samples \cite{krispel2014rules}. These computational frameworks, which encompass a diverse array of methodologies, have witnessed significant evolution over time. Early forays into generative modeling leveraged deep neural networks, exemplified by restricted Boltzmann machines (RBMs) \cite{fischer2012introduction} and deep Boltzmann machines (DBMs) \cite{salakhutdinov2009deep}. Recent advancements have introduced a multitude of innovative approaches, including variational autoencoders (VAEs) \cite{odaibo2019tutorial}, autoregressive models \cite{doi:10.1080/07474938.2021.1899504}, normalizing flows \cite{izmailov2020semi}, generative adversarial networks (GANs) \cite{pan2019recent}, and diffusion models \cite{luo2022understanding}. Notably, the ProGAN/StyleGAN \cite{karras2019style} family has demonstrated remarkable capabilities in producing photorealistic images, predominantly focusing on single-class generation tasks. The emergence of these sophisticated generative techniques has spurred investigations into forensic methodologies geared towards discerning synthetic imagery from authentic counterparts. Particularly noteworthy are recent strides in diffusion models, which have showcased unprecedented proficiency in generating images from textual descriptions \cite{balaji2022ediff}. 

In this paper, we leverage the CIFAKE \cite{CIFAKE} dataset, which harnesses Stable Diffusion \cite{rombach2022high} to generate synthetic images. CIFAKE serves as a valuable resource for training and evaluating AI-generated image detection models, as it provides a diverse collection of synthetic images across various domains. As the field continues to push the boundaries of generative capabilities, the challenge of effectively distinguishing between real and synthetic content grows increasingly complex.

\subsection{AI-Generated Image Recognition}

The identification of manipulated images has a longstanding history in media forensics \cite{4806202}, with established methodologies relying on signals such as resampling artifacts \cite{resampling_article}, JPEG quantization \cite{8267641}, shadows \cite{shadow}, and the detection of operations like image splicing \cite{is} or Photoshop warps \cite{photoshop_warp}. With the proliferation of deep generative methods, particularly in the context of GAN-based techniques \cite{GAN}, recent investigations have delved into the efficacy of discriminative methods for detecting synthesized content. A central inquiry pertains to the generalizability of detectors to unseen methods, with studies indicating that a classifier trained on one GAN model can generalize to others, especially under aggressive augmentations.\cite{deep_discriminative}

Despite successes, challenges emerge when adapting detectors to new generators, where observed high average precision is juxtaposed with low accuracy, indicating proficient separation between real and fake classes but suboptimal calibration. Various techniques, including the utilization of frequency cues \cite{pmlr-v119-frank20a}, co-occurrence matrices \cite{matrix}, pretrained CLIP features\cite{CLIP}, and augmentation with representation mixing \cite{mixing}, have demonstrated effectiveness \cite{dms}. Notably, Ojha et al. \cite{Ojha_23} demonstrate that a simple nearest neighbors classifier improve accuracy, though at the cost of inference time. We expand upon the common observation that even rudimentary classifiers possess a capacity for generalization across various data generators. This exploration involves analyzing and defining their performance within an online context.

Recent investigations into diffusion methods reveal that, contrary to GAN-based detectors' limitations in generalization, diffusion models are detectable and exhibit some degree of mutual generalizability \cite{Ojha_23}. David C. Epstein et al. \cite{epstein2023online} take these studies
further by training a detector on 14 methods in an online
fashion, simulating their release dates, and releasing an accompanying dataset of 570k images. While these works detect whole images, local prediction also offers important use-cases. For instance, in forensic analysis, there's a growing need to identify alterations made by conventional editing tools like Photoshop, such as image warping and splicing\cite{photoshop_warp}. Chai
et al. \cite{Chai_20}  show that patch-based classifiers can generate
heatmaps for regions that contain more detectable cues. We aim to determine whether we can localize inpainted regions. Remarkably, even in the absence of direct access to inpainted examples, employing CutMix augmentation\cite{Yun_2019_ICCV} enables us to utilize entire images effectively for pixel-level predictions.

\section{Proposed Method}
\label{sec:proposed_method}

\subsection{Overview}

\begin{figure}[t!]
    \centering
    \includegraphics[scale=0.3]{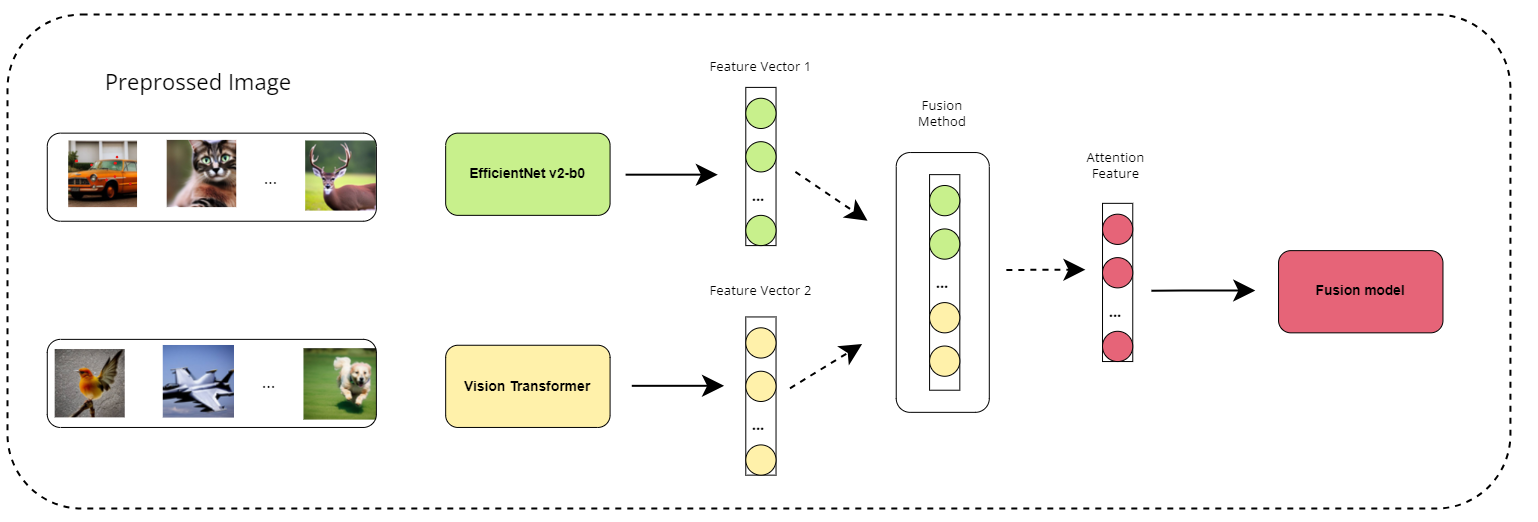}
     \caption{Illustrating of our fusion method between efficientnetv2-b0 model and ViT-b16 model.}
    \label{fig:fusion str}
\end{figure}

Our new approach focuses on using fusion strategies to significantly enhance the accuracy of our AI-generated image recognition model. By combining the strengths of CNNs and (ViTs) through fusion, we aim to create a model that efficient in noticing both tiny local details and broader global context within images. This fusion not only deals with the different types of AI-generated images but also makes sure that CNNs, which are good at pulling out details, and ViTs, which understand the whole picture, work well together. The fusion strategies play a vital role in improving the model's precision and adaptability, making it a potent solution for the task of accurately detecting AI-generated images. Our proposed method is outlined in \autoref{fig:fusion str}.

\subsection{CNN Branch}
In the CNNs branch, we decided to use EfficientNet v2 as the main model of CNNs family to fuse with the one in ViTs family. We choose this model because EfficientNet is a new family of CNNs that has better training speed and better parameter efficiency than previous CNNs models \cite{EfficentNet}. 


\subsection{ViT Branch}
In the ViT branch, our contribution unfolds as we meticulously process input images through a Vision Transformer (ViT) model, leveraging the strengths of the pre-trained "dima806/ai\_vs\_real\_image\_detection" architecture.

To adapt the images to the ViT model's requirements, we apply a series of transformations, including resizing, rotation, sharpness adjustment, and normalization. These transformations ensure that the input images are aligned with the RGB format expected by the ViT model.

The subsequent step involves the creation of a ViTForImageClassification model from the pre-trained checkpoint. This model is configured to map class labels to their corresponding indices, facilitating subsequent model evaluation and interpretation. The number of trainable parameters is calculated, providing insights into the model's complexity and capacity to capture intricate features.

Importantly, the ViT branch's contribution extends beyond mere feature extraction. It involves the orchestration of the entire pipeline, from meticulous image processing to the instantiation of a ViT model capable of distilling global features. These global-aware features, obtained through multi-head self-attention and an MLP, play a pivotal role in our fusion strategy.

During the fusion process with the CNN branch, these features are seamlessly integrated, fostering a comprehensive representation that harnesses both local and global feature extraction mechanisms. This holistic approach amplifies the model's accuracy, resilience, and effectiveness in detecting AI-generated images.
\subsection{Fusion strategy}
The fusion strategy in our proposed method plays a pivotal role in
combining the strengths of the CNN and Vision Transformer (ViT)
branches while ensuring that both spatial and global context information are effectively integrated. In this section, we introduce various feature fusion techniques, including concatenation and linear combination fusion to effectively integrate information from CNNs and ViTs.

\subsubsection{Concatenation}

Our first approach is fusing CNNs and ViTs using concatenation method. This fusion aims to capitalize on the localized feature extraction capabilities of CNNs, which are known for capturing intricate spatial hierarchies, and the global context understanding of ViTs, adept at discerning long-range dependencies within images.

The method involves training dedicated CNN and ViT models for image feature extraction, followed by the extraction of representative features from their intermediate layers. These features are then fused using concatenation or merging techniques, facilitated by a fusion layer. This innovative amalgamation of features creates a unified representation that leverages the complementary advantages of both architectures.

Let \( X_{CNN}\) be the output features from the CNN model, and \( X_{ViT}\)be the output features from the ViT model. The concatenation operation can be represented as follows: \[X_{concatenated} = [X_{CNN},X_{ViTs}], \:\:\:\: (1) \] \[X_{final} = FullyConnectedLayer(X_{concatenated}). \:\:\:(2)\] 

By integrating a classification head onto the fused features, the model becomes proficient in making accurate predictions, thereby offering a robust solution for AI-generated image detection. This method underscores the importance of combining diverse neural network architectures to enhance the overall performance and adaptability of computer vision models. Fine-tuning and customization allow for the optimization of the method according to specific datasets and detection tasks.

\subsubsection{Linear Combination Fusion}

Fusing Convolutional Neural Networks (CNNs) and Vision Transformers (ViTs) linearly involves combining the features learned by both architectures in a linear manner. This approach aims to leverage the strengths of both CNNs, known for their spatial hierarchies through convolutions, and ViTs, which capture long-range dependencies through self-attention mechanisms

Let \(X_{CNN}\) be the feature representation from the CNN architecture, \( X_{ViT} \) be the feature representation from the ViT architecture, \(W_{CNN}\) and \(W_{ViT}\) be the learnable weights for the linear transformations of the CNN and ViT features, respectively, \(b\) be the bias term, \(Y\) is the final result. The linear fusion equation can be written as:

\[Y=W_{CNN}\cdot X_{CNN} + W_{ViT}\cdot X_{ViT} + b. \:\:\:(3)\]

The equation performs a linear combination of the features from the CNN and ViT architectures. The feature representations from both architectures are multiplied by their respective learnable weights (\(W_{CNN}\) and \(W_{ViT}\)), and the results are summed together. This process allows the model to blend information from both architectures linearly.

The addition of the bias term (\(b\)) provides further flexibility in adjusting the combined features.

The resulting fused feature representation \(Y\) is the output of the linear fusion process. This combined representation integrates information from both the CNN and ViT architectures, potentially capturing complementary aspects of the input data.

\section{Experimental Results}
\label{sec:experiments}

\subsubsection{Experiment Settings}

We performed all experiments using the TensorFlow framework on an Ubuntu system. Our experiments were run on two Nvidia T4 GPUs with 16GB of memory each. We selected Binary cross-entropy loss as our loss function for two-label classification. We employed the AdamW optimizer for training our CNN models. The optimizer's weight decay was set to 100 for every CNN, ensuring regularization during training. A momentum of 0.9 was utilized to facilitate faster convergence. Our training process incorporates an early stopping mechanism, stopping training if the validation loss does not improve by a margin of 1000 within five epochs. Additionally, to safeguard model progress, we implement a checkpoint system, preserving the model’s state each time the validation loss experiences reduction of at least 10000.

\subsection{Datasets}
We used the CIFAKE dataset \cite{CIFAKE} as a benchmark to evaluate the proposed network. The dataset contains two classes - "real" and "fake". For "real", the authors collected the images from Krizhevsky and Hinton's CIFAR-10 dataset \cite{CIFAR}. The "fake" class comprises synthetic images generated using Stable Diffusion version 1.4. 
\subsection{Results}
We performed qualitative and quantitative experiments for evaluation. The prediction from our proposed method is compared with the state-of-the-art CNN-based and fused CNN-Transformer networks. We chose the Efficientnet, Resnet50, VGG16, Mobilenet as the pure CNN architectures to employ the evaluation and comparison with our fused model which combines Efficientnet and ViT using our proposed fusion strategies.

\subsubsection{Pure CNNs models performance}

Based on our experimental findings detailed in \autoref{tab:cnn_base_model_result}, the efficiency of various CNN models was evaluated, with EfficientNet v2 emerging as the top performer, boasting an impressive accuracy rate of 97.17\%. Furthermore, in our training process, EfficientNet consistently exhibits stability and high accuracy across the initial epochs of training. Given its consistently superior performance across multiple tests, it stands out as the optimal candidate for combining with Vision Transformer (ViT) models to potentially yield an even more accurate hybrid model.

\begin{table}[t!]
    \centering
    \caption{Result of pure CNNs base model and pure ViTs base model.}
    \label{tab:cnn_base_model_result}
    \begin{tabular}{| c | c | c | c | c | c |} \hline
        \textbf{Model} & \textbf{Loss} & \textbf{Accuracy} & \textbf{Precision} & \textbf{Recall} \\ \hline
         efficientnetv2-b0 & \textbf{10.95\%} & \textbf{97.17}\% & 96.63\% & \textbf{97.74\%} \\ \hline
        Resnet50 & 16.47\% & 95.73\% & \textbf{96.70\%} & 94.69\% \\ \hline
        VGG16 & 12.66\% & 96.28\% & 96.12\% & 96.44\% \\ \hline
        MobilenetV3small & 14.64\% & 94.98\% & 94.53\% & 95.49\% \\ \hline 
        ViT-b16 & 41.60\% & 87.48\% & 88.67\% & 85.94\% \\ \hline
    \end{tabular}
\end{table}

\subsubsection{Pure ViT model performance}
In our experimental evaluation, the ViT model exhibited the least accuracy among the tested models, achieving a score of 87.48\% in \autoref{tab:cnn_base_model_result}. Recognizing this limitation, we expect the fusion technique to improve the performance of the ViT model and also to use its strengths to enhance the performance of the EfficientNet model.

\subsubsection{Fusion result}


Our proposed fusion strategies have successfully achieved the two highest accuracy scores when compared with the accuracy of the two individual models, CNN and ViT. As depicted in \autoref{tab:fusion_feature_result}, the concatenation method achieved the highest accuracy (97.44\%), representing an increase of 0.37\% compared to EfficientNet and 9.96\% compared to ViT. Additionally, the Linear Combination method surpassed our initial expectations during training, achieving an accuracy of 97.32\%. This achievement was realized through adjustments of the weight constants, assigning a weight of 0.6 to EfficientNet and 0.4 to ViTs.



\begin{table}[t!]
    \centering
    \caption{Results of fusion algorithms between efficientnetv2-b0 and ViT-B16.}
    \label{tab:fusion_feature_result}
    \begin{tabular}{| c | c | c | c | c |} \hline
        \textbf{Model} & \textbf{Loss} & \textbf{Accuracy} & \textbf{Precision} & \textbf{Recall} \\ \hline
        Concatenation & \multirow{2}{*}{\textbf{10.74\%}}  & \multirow{2}{*}{\textbf{97.44\%}} & \multirow{2}{*}{97.60\%} & \multirow{2}{*}{\textbf{97.27\%}}\\
        (efficientnet + ViT) & & & & \\ \hline
        Linear Combination & \multirow{2}{*}{11.24\%}  & \multirow{2}{*}{97.32\%} & \multirow{2}{*}{\textbf{98.02\%}} & \multirow{2}{*}{96.59\%}\\
        (efficientnet + ViT) & & & & \\ \hline
    \end{tabular}
\end{table}

\subsubsection{Reduce brightness by 50\%}

After applying the fusion method, the accuracy slightly increases that of the base model. Thus, we attempt to assess our model's performance under challenging conditions by reducing the brightness of testing images.

In this experiment, we convert the image color space from RGB-base to HSV-base and then decrease the "Value" parameter by 50\% in the validation dataset, effectively reducing image brightness by half \cite{stackoverflowFastChange}. Then we use our pretrained CNNs base models and our proposed model to evaluate that modified validation dataset. The performances of the CNNs base models and our custom models described in \autoref{tab:reduce_brightness_result_CNNs} and \autoref{tab:reduce_brightness_result_Fusions} respectively.

\begin{table}[t!]
    \caption{Results of pure CNNs base model and pure ViTs base model on the reduce brightness dataset.}
    \label{tab:reduce_brightness_result_CNNs}
    \centering
    \begin{tabular}{| c | c | c | c | c | c |} \hline
        \textbf{Model} & \textbf{Loss} & \textbf{Accuracy} & \textbf{Accuracy drop} \\ \hline
        ResNet50 & 33.55\% & 86.59\% & 9.14\% \\ \hline
        {VGG16} & \textbf{18.75\%} & \textbf{94.78\%} & \textbf{1.5\%} \\ \hline
         {efficientnetv2-b0} & 49.49\% & 81.67\% & 15.50\% \\ \hline
         {MobilenetV3small} & 69.36\% & 50\% & 44.98\% \\ \hline
         {ViT-b16} & 62.91\% & 68.04\% & 19.44\%\\ \hline
    \end{tabular}
    
\end{table}

\begin{table}[t!]
    \centering
    \caption{Resulst of fusion algorithms on the reduce brightness dataset.}
    \label{tab:reduce_brightness_result_Fusions}
    \begin{tabular}{| c | c | c | c | c | c |} \hline
        \textbf{Model} & \textbf{Loss} & \textbf{Accuracy} & \textbf{Precision} & \textbf{Recall} \\ \hline
        Linear Combination & \multirow{2}{*}{28.23\%}  & \multirow{2}{*}{91.62\%} & \multirow{2}{*}{90.30\%} & \multirow{2}{*}{93.26\%}\\
        (efficientnet + ViT) & & & & \\ \hline
        Linear Combination & \multirow{2}{*}{\textbf{15.62\%}}  & \multirow{2}{*}{\textbf{95.29\%}} & \multirow{2}{*}{\textbf{93.08\%}} & \multirow{2}{*}{\textbf{97.85\%}}\\
        (VGG + ViT) & & & & \\ \hline
    \end{tabular}
    
\end{table}

The observation that most CNN base models experience a sharp decline in accuracy when faced with images of reduced brightness underscores the complexity of the task at hand. Notably, the VGG16 model stands out as being relatively resilient to such challenges, with only a marginal 1.5\% drop in accuracy. Leveraging this insight, we devised our custom fusion model, which not only preserves the robustness of VGG16 but also harnesses the advanced capabilities of ViT-b16. The result is a substantial enhancement in accuracy, culminating in an impressive 95.29\% accuracy rate on the processed validation dataset with challenging brightness conditions.

Undoubtedly, our method has made significant strides in enhancing the accuracy of detecting AI-generated images. By combining the strengths of the VGG16 and ViT-b16 models through a fusion approach, we've achieved a noteworthy improvement in accuracy metrics. This improvement is particularly pronounced when confronted with images subjected to lower brightness conditions, a scenario where conventional CNN base models often struggle.

This remarkable improvement underscores the effectiveness of our approach in enhancing the model's ability to discern AI-generated images with greater precision and reliability. By leveraging the complementary strengths of different architectures and intelligently fusing them, we've unlocked new levels of performance, paving the way for more accurate and reliable detection of AI-generated content in various real-world scenarios.

\section{Conclusion}
\label{sec:conclusion}
Our study introduces a novel method for recognizing AI-generated images, aiming to enhance prediction efficiency and accuracy across several popular models. At the heart of our proposed approach lies the fusion of Convolutional Neural Networks (CNNs) and Vision Transformer architectures. We explore two fusion strategies—concatenation and linear combination—which yield slight accuracy improvements compared to using the models separately.

Our methodology begins by extracting feature vectors from both EfficientNet and Vision Transformer models. These vectors are then combined into a unified output vector using mathematical formulas and algorithms. This fusion process enables the models to leverage the strengths of both CNNs and Vision Transformer, resulting in more robust predictions.

To validate the effectiveness of our approach, we subjected the testing dataset to challenging conditions. Remarkably, our experiments reveal that the fusion of VGG16 and ViT achieved the highest accuracy under these demanding circumstances. This finding underscores the resilience and effectiveness of our fusion technique, particularly when faced with complex and varied image data.

Overall, our experiments demonstrate that our proposed fusion technique significantly enhances feature extraction accuracy and image recognition capabilities compared to individual branch models. By seamlessly integrating CNNs and Vision Transformer architectures, we pave the way for more accurate and efficient AI-generated image recognition systems.

\section*{Acknowledgement}
	
This research is supported by research funding from Faculty of Information Technology, University of Science, Vietnam National University - Ho Chi Minh City.

\bibliographystyle{splncs04}
\bibliography{reference}

\end{document}